\documentclass[10pt,twocolumn,letterpaper]{article}

\usepackage[english]{babel}
\usepackage{times}
\usepackage{epsfig}
\usepackage{subfig}
\usepackage{graphicx}
\usepackage{amsmath}
\usepackage{amssymb}
\usepackage{algorithm}
\usepackage{algpseudocode}
\usepackage{colortbl}
\usepackage{multirow}
\usepackage{xcolor}
\usepackage{booktabs}
\usepackage{titling}
\usepackage{listings}
\definecolor{mygreen}{rgb}{0.0,0.7,0.3}
\definecolor{myblue}{rgb}{0.0,0.5,1.0}
\usepackage[T1]{fontenc}
\usepackage[pagebackref=true,breaklinks=true,letterpaper=true,colorlinks,bookmarks=false]{hyperref}
\title{CLIP-ReIdent: Contrastive Training for Player Re-Identification}
\author{Konrad Habel \hspace{4em} Fabian Deuser \hspace{4em} Norbert Oswald \\
University of the  Bundeswehr Munich \\
Institute for Distributed Intelligent Systems \\
Munich, Germany \\
{\tt\small konrad.habel@unibw.de, fabian.deuser@unibw.de, norbert.oswald@unibw.de}
}
\begin{document}
\date{}
\maketitle
\begin{abstract}
\textit{
Sports analytics benefits from recent advances in machine learning providing a competitive advantage for teams or individuals. One important task in this context is the performance measurement of individual players to provide reports and log files for subsequent analysis. During sport events like basketball, this involves the re-identification of players during a match either from multiple camera viewpoints or from a single camera viewpoint at different times. In this work, we investigate whether it is possible to transfer the out-standing zero-shot performance of pre-trained CLIP models to the domain of player re-identification. For this purpose we reformulate the contrastive language-to-image pre-training approach from CLIP to a contrastive image-to-image training approach using the InfoNCE loss as training objective. Unlike previous work, our approach is entirely class-agnostic and benefits from large-scale pre-training. With a fine-tuned CLIP ViT-L/14 model we achieve 98.44\% mAP on the MMSports 2022 Player Re-Identification challenge. Furthermore we show that the CLIP Vision Transformers have already strong OCR capabilities to identify useful player features like shirt numbers in a zero-shot manner without any fine-tuning on the dataset. By applying the Score-CAM algorithm we visualise the most important image regions that our fine-tuned model identifies when calculating the similarity score between two images of a player.}
\end{abstract}

\section{Introduction}
The performance of person re-identification has been greatly improved in recent years by advances in deep learning. This has led to a saturation on standard benchmarks~\cite{Li2014DeepReid, Zheng2015ScalableReid, Ristani2016PerformanceMA} for pedestrian identification. Due to the setting of these datasets recent approaches~\cite{wieczorek2021unreasonable,fu2021unsupervised,he2021dense} focused on multi-view dependent features, cloth-changing~\cite{qian2020long, gu2022clothes} and in-the-wild identification. This differs in parts from player re-identification, therefore further investigation is needed. Sport analytics profit tremendously from player tracking combined with re-identification~\cite{Cioppa2022SoccerNetTracking, Vats2021PlayerTA} to analyse individual performance of a player. 

What distinguishes visual player re-identification from pedestrian identification are on the one hand homogeneous backgrounds, since e.g. basketball courts or ice hockey arenas are constructed in standardised design. On the other hand, all players on a team wear identical shorts and jerseys, which means that features such as the player's shirt number, shoes, or face become more important for recognising a particular player of a team. The images are extracted by using bounding boxes of the players. Therefore the images of a player are often of low resolution and may contain motion blur. Facial recognition would be useful, but is difficult in most of these images. Due to the nature of team-based sports the back number is the best feature to identify a particular player. Therefore, optical character recognition (OCR) can further support identification through jersey number recognition.

In this paper, we utilise the Contrastive Language-Image Pre-training (CLIP)~\cite{radford2021learning} and reformulate it as contrastive image representation learning to determine the distance between images. The main contributions are:
\begin{enumerate}
    \item We reformulate the CLIP objective for single modal class-agnostic re-identification and retrieval.
    \item An extensive analysis of the zero shot capabilities and region importance of our approach.
\end{enumerate}

\section{Related Work}
Previous approaches mainly focused on jersey number recognition to apply an identification based on the detected numbers~\cite{Gerke2015Soccer, Li2018Jersey, Liu2019PoseGuidedRF, Nady2021PlayerII} or on part-based classification~\cite{Senocak2018cnnsvm}.

Many datasets in player re-identification are private and use tracklets for identification. Tracklets are extracted consecutive frames from a video, only displaying the bounding box of the detected player. Vats et al.~\cite{Vats2021PlayerTA} used 1D convolutions on top of a ResNet-18~\cite{he2016deep} to predict jersey numbers from ice hockey players. Each extracted frame is encoded by a Convolutional Neural Network (CNN) as a vector and the 1D convolutions are then applied to this sequence. The dataset is unbalanced and if no jersey number is visible the sample is mapped to a $null$ class. Therefore, this approach is limited to predefined classes and also requires a team identifier to predict the identity of the currently tracked player.

To address the problems of fixed classes and reliance on jersey numbers Teket et al.~\cite{TeketBasketball2020} proposed a real-time capable pipeline for player detection and identification. For their approach, they use a triplet loss~\cite{hoffer2015deep, chechik2010large} with a Siamese network~\cite{koch2015siamese} to separate players from each other. Through unsupervised training, class-agnostic player detection is possible. They also showed that a classification loss on the 22 players in their dataset performs worse than the triplet loss.

Due to the lack of public datasets, Maglo et al.~\cite{maglo2022efficient} proposed a semi-interactive system for player re-identification. It allows efficient learning based on a small number of annotations and uses a transformer-based architecture for recognition. Multiple tracklets are encoded by a CNN, and a transformer encoder-decoder is used to predict the player ID. A class-based ID loss (cross-entropy) as well as a triplet loss which is applied to the embeddings of the tracklets after the decoder are used for this purpose.

A similar architecture is used by Vats et al.~\cite{vats2021ice} to predict the jersey number as classification. The cross-entropy loss for jersey numbers is calculated for two-digits as well as for all single digits. 

In their analysis~\cite{comandur2022sports} Comandur et al. identified hierarchical data sampling as beneficial. Similar actions and images from the same match are sampled into a batch for training. They also introduced a centroid loss to increase the margin between clusters. This loss is then combined with a triplet loss and a class-based ID loss (cross-entropy).
An approach similar to ours by Fu et al.~\cite{fu2021unsupervised} uses MoCo~\cite{he2020momentum} to learn person dependent features in a unsupervised manner. However, through the single-view learning in the MoCo approach multi-view features are not learned.

\section{Methodology}
We propose a new method to use a pre-trained CLIP image encoder as a Siamese network. In addition a contrastive training objective inspired by previous work in image-text matching~\cite{oord2018representation, Zhang2020ContrastiveLO} is used. To the best of our knowledge we are the first who provide a novel framework in using CLIP for Re-Identification.

\subsection{Contrastive Pre-training} \label{methodology_clip}

\begin{figure}[]
\centering
	\includegraphics[width=0.45\textwidth]{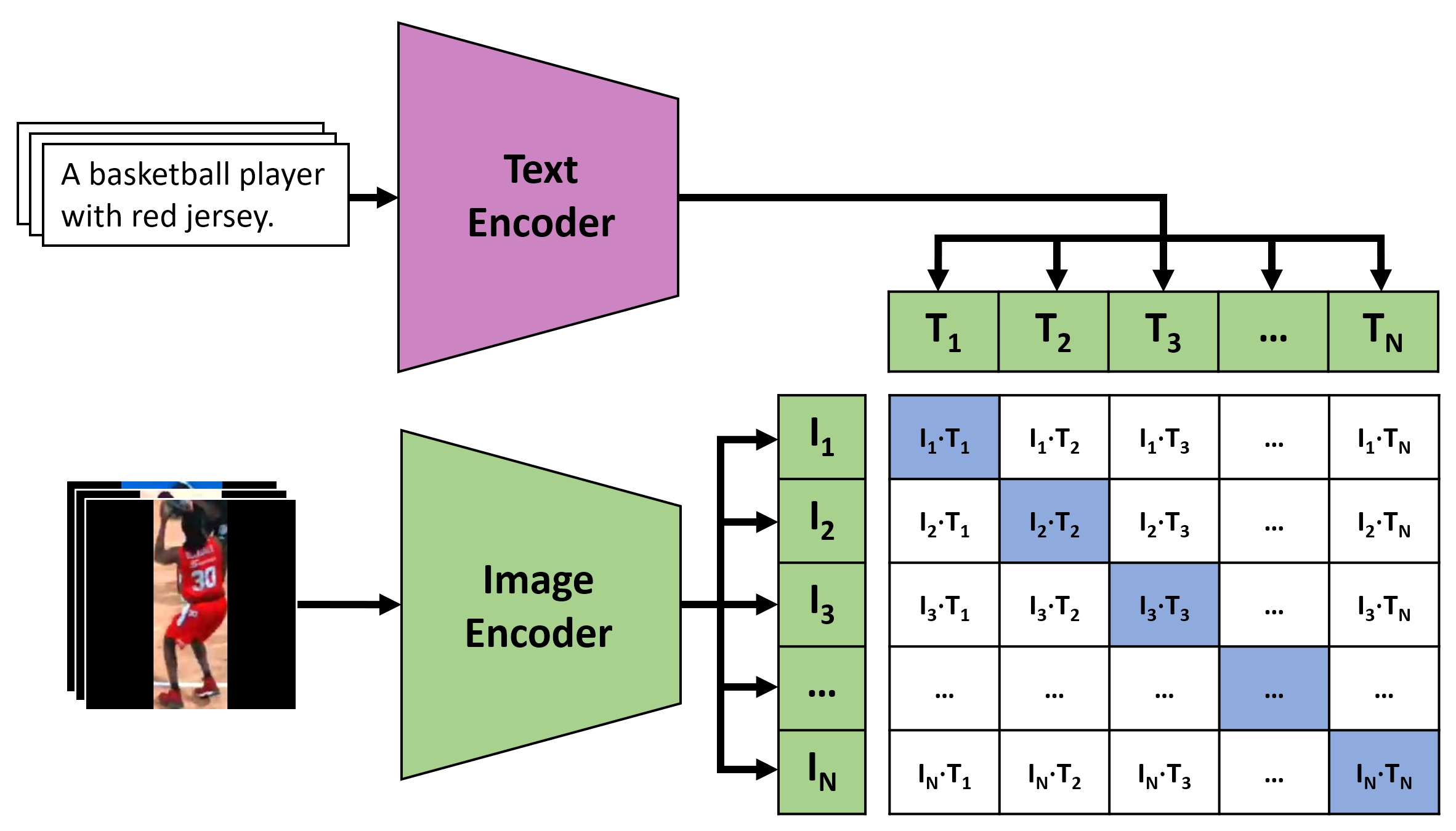}
	\caption{Multimodal approach: Training procedure of CLIP matching image-text pairs to maximise the cosine similarity between pairs within a batch.}
	\label{img:clip_org}
\end{figure}

All CLIP models from OpenAI are trained on 400 million image-text pairs, with the objective to match a given caption to an image and vice versa. The image-text pairs are crawled from web sources based on $500.000$ queries to cover many concepts during training. During the contrastive pre-training the images are encoded by an image encoder like Vision Transformer (ViT)~\cite{dosovitskiy2020image} or ResNet and the texts are encoded by a Transformer~\cite{vaswani2017attention}. During training the cosine similarity between the corresponding pairs in a batch is maximised while the cosine similarity to all other pairs is minimised, as shown in Figure~\ref{img:clip_org}. This enables zero-shot capabilities because novel classes or concepts can be retrieved by matching an image and multiple text prompts by the highest cosine similarity.

In contrast to a model pre-trained on ImageNet~\cite{Russakovsky2015Imagenet}, which produces an output for a fixed number of classes, CLIP models generate an embedding vector of an image with a size depending on the used architecture.

\subsection{Approach for Re-Identification} \label{methodology_approach}
\begin{figure}[]
\centering
	\includegraphics[width=0.45\textwidth]{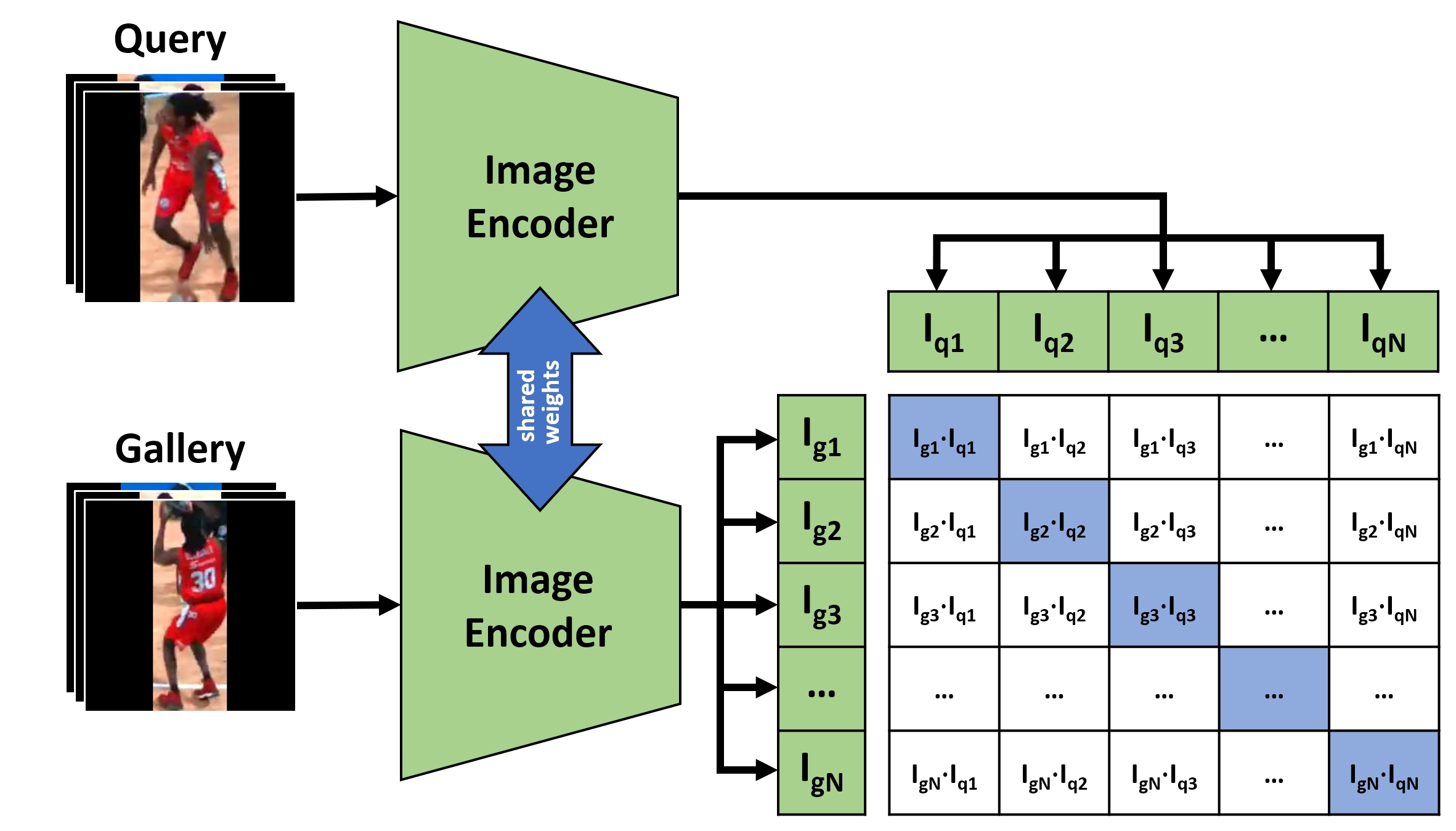}
	\caption{Vision-only approach: Our training procedure matching image-image pairs and shared parameters of the image encoder.}
	\label{img:clip}
\end{figure}

We reformulate the multimodal approach from CLIP for the calculation of similarity between image-text pairs into a vision only approach by calculating the similarity of image-image pairs. As shown in Figure~\ref{img:clip}, two images of the same player are entered into the network with the aim to generate embeddings with a small distance for images with high similarity. This approach can also be used with models pre-trained on ImageNet. In that case, we use as image embedding the output of the model before the last classification layer.

To minimise the distance between the embeddings of the query image and the gallery image, we use the InfoNCE loss~\cite{Sohn2016multiclassnpair} during training. Similar to the work of Chen et al.~\cite{chen2020simple} and He et al.~\cite{he2020momentum} a visual representation is learned. However, in contrast we do not use augmented instances of the same image. Instead we have tracklets for each player with images of different poses and camera angles. The objective is to minimise the distance between images of the same tracklet.

While the triplet loss always samples a positive and a negative sample, the InfoNCE loss is calculated over one positive versus $n-1$ negative samples, where $n$ is the number of samples in the batch. Oord et al.~\cite{oord2018representation} were able to show that for a larger $n$ and thus more negative samples, the lower bound on mutual information between positive samples is maximised.

\lstset{
language=Python,
frame=lines,
commentstyle=\color{mygreen},
label={lst:code_direct},
caption={Numpy-like pseudo code for the reformulated loss function for a vision only image to image matching.},
basicstyle=\footnotesize\ttfamily
}
\begin{lstlisting}[language=Python, float=h]
# image_encoder   - Vision Transformer or CNN
# I_q[n, h, w, c] - minibatch of query images
# I_g[n, h, w, c] - minibatch of gallery images
# n               - batch size
# t               - learned temperature parameter

# extract feature representations [n, embedding_dim]
I    = np.concatenate([I_q, I_g], axis=0)
I_qg = image_encoder(I)
I_qg = l2_normalise(I_qg, axis=1)
I_q  = I_qg[:n, :]
I_g  = I_qg[n:, :]

# scaled pairwise cosine similarities [n, n]
logits_q = np.exp(t) * np.dot(I_q, I_g.T) 
logits_g = logits_q.T

# symmetric loss function
labels = np.arange(n)
loss_q = cross_entropy_loss(logits_q, labels)
loss_g = cross_entropy_loss(logits_g, labels)
loss   = (loss_q + loss_g)/2

\end{lstlisting}

For the MMSports 2022 Player Re-Identification challenge however, this is an issue because the number of different players is relatively small. Therefore, we use a batch size of $n=16$ with a custom batch sampler. A pair of gallery and query images of the same player represents a unique instance during training. Our sampler takes care that only instances of different players are in the batch. Otherwise, this would add noise to the model, as no clear assignment of a gallery image to the corresponding query image of the player would be possible within the batch. This maximises the cosine similarity within the instance while minimising the similarity to all other player instances in the batch. However, our sampler does not take into account that a player may also be visible on images of an instance from another player in the same batch. To mitigate this problem, we use label smoothing, which softens the training objective and leads to higher mAP scores. Because we have pairs of images and each pair is encoded by the same image encoder, we can encode both query and gallery images of the same batch at once. This doubles the effective batch size.

Following recent work~\cite{quispe2021top, Gong2021EliminateDW, zang2021learning} we use re-ranking~\cite{zhong2017re} as post-processing step for the final distance matrix, but also provide mAP scores without re-ranking.

\subsection{Zero-shot Capabilities} \label{methodology_Zero-shot}
Zero-shot learning aims to learn classes only based on descriptions~\cite{larochelle2008zero}, thus allowing to retrieve novel concepts during inference. Based on the strong zero-shot capabilities of CLIP further investigation of the capabilities and their implication about learned concepts is necessary. For this evaluation we annotated 100 images of different players from the train set, where the jersey number is visible. We use for our ablation study the following attributes: jersey number, jersey colour, sex and skin colour. CLIP is trained with multimodal data to include textual information in the image embedding space and vice versa. This allows us to perform a zero-shot classification based on text prompts for the annotated attributes. The text prompts are embedded by the CLIP text encoder and the predicted class is identified based on the closest cosine similarity between image and text-prompts. We did not further fine-tune CLIP to determine the zero-shot capabilities on this out of domain data. For example the classes for the jersey colour prediction are encoded text prompts with the template: {\itshape "a $\lbrace$c$\rbrace$ jersey, colour $\lbrace$c$\rbrace$"} with $c \in [black, blue, green, orange, red, white, yellow]$. The embedding of the sentence with the highest cosine similarity to the image embedding is our predicted class in the zero-shot setting.

Furthermore, we leverage the Score-CAM algorithm~\cite{wang2020score} to demonstrate the ability of CLIP Vision-Transformers to localise the jersey number in a zero-shot scenario and visualise the similarity of query and gallery images with our fine-tuned model. This algorithm works in two phases. First the activation maps of a layer are extracted and up-sampled to the original size of the image. This up-sampled activation maps are then used as mask on the original image. These $M$ masked images, where $M$ is the number of activation maps, are then encoded by the image encoder and based on the prediction result the weights of the activation maps can be calculated. The final result is then a linear combination of these weights and the activation map, resulting in a more accurate visualisation compared to GradCAM~\cite{selvaraju2017grad}.

\begin{table*}[h]
\centering
\caption{Results on the Test-Set (Paper submission) and model details.}
\begin{tabular}{@{}lcc|ccccr@{}}
\toprule
& \multicolumn{2}{c|}{\textbf{mAP [\%]}} & & input  & embedding & max  & \\
Model                     & w/o rerank  & w/ rerank & pre-training & resolution & dimension & learning rate & parameter  \\ 
\midrule 
convnext\_base\_in22ft1k  & $89.9$    & $96.0$   & ImageNet & $224$ & $1024$ & $4\times10^{-4}$ & $87\,m$\\
convnext\_large\_in22ft1k & $91.2$    & $96.9$   & ImageNet & $224$ & $1536$ & $4\times10^{-4}$ & $196\,m$\\
vit\_base\_patch16\_224   & $89.9$    & $95.4$   & ImageNet & $224$ & $768$  & $4\times10^{-5}$ & $85\,m$\\
vit\_large\_patch16\_224  & $92.5$    & $97.1$   & ImageNet & $224$ & $1024$ & $4\times10^{-5}$ & $303\,m$\\
\midrule 
RN50x16                   & $88.5$    & $94.9$   & CLIP    & $384$ & $768$ & $4\times10^{-4}$ & $136\,m$             \\ 
ViT-B/16                  & $95.0$    & $98.1$   & CLIP    & $224$ & $768$ & $4\times10^{-5}$ & $57\,m$              \\ 
ViT-L/14                  & $\textbf{96.9}$  & $\textbf{98.2}$  & CLIP & $224$ & $1024$ & $4\times10^{-5}$ & $202\,m$               \\ 
\bottomrule 
\end{tabular}
\label{tab:results_test}
\end{table*}

\section{Evaluation}
\subsection{Dataset}
The DeepSportRadar Player Re-Identification dataset~\cite{zandycke2022deepsportradarv} provides a given train, test and challenge split with the distribution in Table~\ref{tab:dataset}. The training set consists of 436 different players with mostly 20 images per player.

\begin{table}[H]
\centering
\caption{Distribution of the dataset.}

\begin{tabular}{@{}lrrr@{}}
\toprule

split           & query      & gallery    & avg. size $(h\times w)$  \\ 
\midrule 
train          & $436$       & $8.133$      & $(209\times100)$      \\ 
test           & $50$        & $910$        & $(223\times101)$    \\ 
challenge      & $468$       & $8.703$      & $(208\times102)$  \\ 

\bottomrule 
\end{tabular}
\label{tab:dataset}
\end{table}

The difficulty for a re-identification task lies primarily in the low resolution of the images, which means that facial features can only be recognised inadequately. The average image size of the training dataset is only $209\times100$ pixels. In addition, the jersey number is only visible on a fraction of the images, which makes it much more difficult to distinguish players of the same team with similar stature and skin colour. On some of the images two or even three players are visible. If there is a corresponding query image for both players from a gallery image, it is sometimes difficult even for humans to distinguish to which query this gallery image should be assigned. The only possibility in such cases is to look closely at which player is really completely visible in the image. However, the generation of the bounding box and extraction of the content does not always seem to be correct. Sometimes parts of a players are cut off and not completely in the visible area. 

For the ablation study we additionally annotated 100 images of different players from the training set with the constraint, that the jersey number must be visible. The attributes jersey number, shirt colour, skin colour and sex of the player were annotated by two independent annotators.

\subsection{Experiments and Results}

\subsubsection{Training Details}
\label{train_details}

The CLIP models from OpenAI are pre-trained with a resize method, where the smaller edge of an image is re-scaled to the input size of the network and a quadratic center crop is performed. This method is not suitable for the present dataset, as it would cut off large parts of most of the images. Instead we use zero-padding to circumvent this problem and also preserve the proportions of the players on the images. We only use random horizontal flip as augmentation method. However, this could have a negative effect on the ability of the network to recognise jersey numbers, but when using random horizontal-flip during training the mAP does not decrease on the test set. 

For the fine-tuning of Vision-Transformer based models with the contrastive learning objective, we use  AdamW~\cite{loshchilov2017decoupled} as optimiser with a polynomial learning rate schedule with warm-up, reaching a maximum learning rate of $4\times10^{-5}$ after two epochs and a minimum learning rate of $4\times10^{-6}$ at the end of the training. For CNNs a higher learning rate of $4\times10^{-4}$ at maximum and $4\times10^{-5}$ at the end of the schedule is used. The label smoothing value is set to $0.1$. All models of Table~\ref{tab:results_test} are trained for 8 epochs and the best checkpoint based on the mAP score is selected. As input resolution we use the image size already used for pre-training of the networks, according to Table \ref{tab:results_test}.

\subsubsection{Results on the Test-Set} \label{results_test}
Table~\ref{tab:results_test} shows the results on the test split for different kind of pre-training methods, model size and architecture. We achieve the best results with the CLIP Vision-Transformers pre-trained with a contrastive learning objective. We use from the original CLIP model only the vision encoder and drop the text encoder and all other text related parts like the embedding and projection layers. We also do not use the vision projection layer of the CLIP ViT models. ViT-B/16 uses a projection layer from $768 \rightarrow 512$ and ViT-L/14 a projection layer from $1024 \rightarrow 768$ to match the output size of the vision encoder with the output size of the text encoder.

The ViT-L/14 model with around 202 million parameters for the vision encoder is a relative large model but it delivers from all tested models the best mAP value of 96.9\% without re-ranking and 98.2\% with re-ranking on the test split. Good results can also be achieved with the ViT-B/16 model, that can be considered as a small neural network with around 57 million parameters. In relation to its parameter count and computational costs the ViT-B/16 achieves with 95.0\% mAP without re-ranking and 98.1\% mAP with re-ranking the best performance of all tested models on the test set.

We also tested CNNs from CLIP, but these models perform significantly worse compared to the CLIP ViT models on the re-identification task. However, models pre-trained on ImageNet do not show such large differences in terms of mAP between the CNN and ViT architecture, when fine-tuned with our approach on the dataset. We use two different sized state-of-the-art ConvNext~\cite{liu2022convnet} CNNs and Vision Transformers for the comparison to the CLIP models. These models are also trained with our proposed objective to compare different pre-training strategies. For these models and pre-trained checkpoints we refer to PyTorch Image Models~\cite{rw2019timm}. Neither of these models trained on ImageNet achieve the same performance as the ViT-B/16 or ViT-L/14 from CLIP on the challenge task. The vit\_large\_patch16\_224 Vision Transformer comes closest to the results of the both CLIP Vision Transformers but only with re-ranking and is with 303 million parameter a relative huge model in comparison to the ViT-B/16.
  
\subsubsection{Results on the Challenge-Set} \label{results_challenge}

Table~\ref{tab:results_challenge} shows the results on the challenge set. We trained for the challenge results the two best models on 8569 train + 960 test images with hyperparameters, derived from experiments and results shown in Table~\ref{tab:results_test}. The only difference in the hyperparameters of the two models is the number of epochs, for how long the models are trained. The large ViT-L/14 is trained for only 4 epochs, whereas the ViT-B/16 is trained for 8 epochs. This leads to a mAP of 97.56\% for the ViT-B/16 and mAP of 98.44\% for the larger ViT-L/14 model. Both results are the outcome of a single model without any ensemble but we use per player re-ranking as post-processing step on the generated distance matrix. 

\begin{table}[h]
\centering
\caption{Results on the Challenge-Set (Paper submission).}
\begin{tabular}{@{}lccc@{}}
\toprule
Model     & mAP [\%]         & rank-1 [\%]      & rank-5 [\%]      \\ 
\midrule 
ViT-B/16  & $97.56$          & $98.72$          & $99.36$          \\ 
ViT-L/14  & $\textbf{98.44}$ & $\textbf{99.15}$ & $\textbf{99.79}$ \\ 
\bottomrule 
\end{tabular}
\label{tab:results_challenge}
\end{table}

\subsection{Ablation Study} \label{ablation}

To gain further insight why the CLIP ViT models outperform models pre-trained on ImageNet, we compare how well the models of Table~\ref{tab:results_test} perform on the re-identification task without any fine-tuning. In addition we analyse if CLIP models are able to identify useful attributes of a player without any training on this task. We apply the Score-CAM algorithm in a zero-shot scenario to localise jersey numbers and on our final fine-tuned model to visualise regions of high similarity between query and gallery images. 

\subsubsection{Zero-Shot Re-Identification} \label{ablation_zero_shot_reident}

First we analyse how good different models are in a zero-shot scenario on the re-identification task, without any training on the DeepSportRadar Player Re-identification dataset. For this analysis we use the models from Table~\ref{tab:results_zero_shot} to embed the images of the dataset and evaluate how well these models perform if not explicitly trained on the data.  It can be shown that models, trained with a contrastive pre-training objective, achieve better result in this scenario than models pre-trained on an image classification task on ImageNet. 

\begin{table}[h]
\centering
\caption{Zero-shot results on the Test-Set.}
\begin{tabular}{@{}lcc@{}}
\toprule
& \multicolumn{2}{c}{\textbf{mAP [\%]}} \\
Model                     & w/o rerank  & w/ rerank  \\ 
\midrule 
convnext\_base\_in22ft1k  & $44.8$   & $61.6$  \\
convnext\_large\_in22ft1k & $50.0$   & $66.2$  \\
vit\_base\_patch16\_224   & $43.0$   & $59.7$  \\
vit\_large\_patch16\_224  & $52.9$   & $69.4$  \\
\midrule 
RN50x16                   & $54.8$   & $70.6$  \\ 
ViT-B/16                  & $63.1$   & $81.8$  \\ 
ViT-L/14                  & $\textbf{66.8}$ & $\textbf{85.0}$\\ 
\bottomrule 
\end{tabular}
\label{tab:results_zero_shot}
\end{table}

CLIP models are better in generating useful embeddings for a similarity matching task, due to the contrastive pre-training with millions of image-text pairs. The objective of minimising the distance between image and text representation is better suited as pre-training task than a typical classification task with a fixed number of defined classes. 

\subsubsection{Zero-Shot Attributes Classification} \label{ablation_zero_shot_attributes}

With these findings, we investigate whether the two best ViT models of CLIP are able to extract useful attributes per player that humans would also use when it comes to identifying a particular player. These attributes cover the jersey number, shirt colour, skin colour and sex of the player. For this purpose, we use the self-annotated subset of the training data. Following Buolamwini et al.~\cite{pmlr-v81-buolamwini18a} we use the Fitzpatrick skin types to determine our labels of the skin colour prediction. Type 1 to 3 are labeled as "white" and Type 4 to 6 are labeled as "black".

For the prediction of these four attributes, no training needs to be done when using a CLIP model. Instead we are leveraging the zero-shot capabilities of these models.  We only need to specify for all particular classes of an attribute an appropriate text prompt, that is close to the image representation we are looking for. These text-prompts for every attribute are shown in Table~\ref{tab:zero_shot_text_prompts}. 

Table \ref{tab:results_attributes} shows the results on four different attributes. For our zero-shot classification we report the classification accuracy for each attribute. While sex and skin colour prediction are binary classification the jersey number and jersey colour prediction is a multi class problem. Both CLIP ViT models can astonishingly well identify the jersey number and colour as well as the sex and skin colour of a player with high accuracy, without the need of explicit training.

In line with the findings of Radford et al.~\cite{radford2021learning} our results in Table~\ref{tab:results_attributes} show that the ViT-L/14 has excellent OCR capabilities. The model is able to recognise jersey numbers even without training on the dataset. We also calculate the $top3$ accuracy for the jersey number, because sometimes if numbers consist of two digits, only one digit is recognised as $top1$ answer. As an example, for the number 12 the $top1$ answer can also be either 1 or 2 and the number 12 may only occur among the $top3$ answers. Same for the jersey colour. The jerseys usually have two colours, at least the back number is always in a second colour. It is difficult to create more specific text-prompts to circumvent this shortcoming, because CLIP has not been trained for such a task.

\begin{table}[H]
\centering
\caption{Zero-shot results for attributes on 100 images of different players of the train set.}
\begin{tabular}{@{}lcc@{}}
\toprule
& \multicolumn{2}{c}{\textbf{Accuracy}} \\
Attribute       & ViT-B/16  & ViT-L/14 \\ 
\midrule 
jersey number   & $top1: 0.56$   & $top1: 0.87$ \\
                & $top3: 0.85$   & $top3: 0.99$ \\
\midrule 
jersey colour   & $top1: 0.81$   & $top1: 0.73$ \\
                & $top2: 0.96$   & $top2: 0.90$ \\
\midrule 
sex             & $top1: 0.90$   & $top1: 0.93$ \\
\midrule 
skin colour     & $top1: 0.83$   & $top1: 0.87$ \\

\bottomrule 
\end{tabular}
\label{tab:results_attributes}
\end{table}

We only annotated images where the back of the player is visible, therefore it is even more challenging to determine the sex of the player. Nevertheless, CLIP models are relatively good at identifying whether the player is male or female. 

Due to the imbalance of the instances per attribute among the 100 labelled images of different players of the train set, we provide in addition to the results in Table~\ref{tab:results_attributes} also the unweighted average recall, precision and F1 scores for these attributes in Table~\ref{tab:recal_results_attributes}. 

\begin{table}[hbt!]
\centering
\caption{Results of unweighted mean (macro average) for attributes on 100 images of different players of the train set.}
\begin{tabular}{@{}lrrrrr@{}}
\toprule
Attribute     & Model     & Precision    & Recall       & F1-score     \\ 
\midrule 
jersey number & ViT-B/16  & $0.399$     & $0.446$     & $0.379$     \\ 
              & ViT-L/14  & $0.796$     & $0.821$     & $0.795$     \\ 
\midrule 
jersey colour & ViT-B/16  & $0.773$     & $0.884$     & $0.812$     \\ 
              & ViT-L/14  & $0.770$     & $0.769$     & $0.693$     \\ 
\midrule 
sex           & ViT-B/16  & $0.823$     & $0.823$     & $0.823$     \\ 
              & ViT-L/14  & $0.863$     & $0.911$     & $0.884$     \\ 
\midrule 
skin colour   & ViT-B/16  & $0.828$     & $0.848$     & $0.827$     \\ 
              & ViT-L/14  & $0.861$     & $0.865$     & $0.863$     \\ 
              
\bottomrule 
\end{tabular}
\label{tab:recal_results_attributes}
\end{table}

\begin{table*}[hbt!]
\centering
\caption{Text prompts used for attribute classification.}
\begin{tabular}{@{}lll@{}}
\toprule
Attribute       & Text-Prompt    & Classes: c\\ 
\midrule 
jersey number:   & "a basketball player with jersey number $\lbrace$c$\rbrace$"   & $[1,2,...,32]$ \\
jersey colour:   & "a $\lbrace$c$\rbrace$ jersey, $\lbrace$c$\rbrace$ colour"   &  $[$black, blue, green, orange, red, white, yellow$]$ \\
sex:             & "a $\lbrace$c$\rbrace$ basketball player"   & $[$male, female$]$ \\
skin colour:     & "a $\lbrace$c$\rbrace$ basketball player"   & $[$white, black$]$ \\
\bottomrule 
\end{tabular}

\label{tab:zero_shot_text_prompts}
\end{table*}

\subsubsection{Zero-Shot Jersey Number Localisation} \label{ablation_zero_shot_localisation}

\begin{figure}[H]
\centering
	\includegraphics[width=0.45\textwidth]{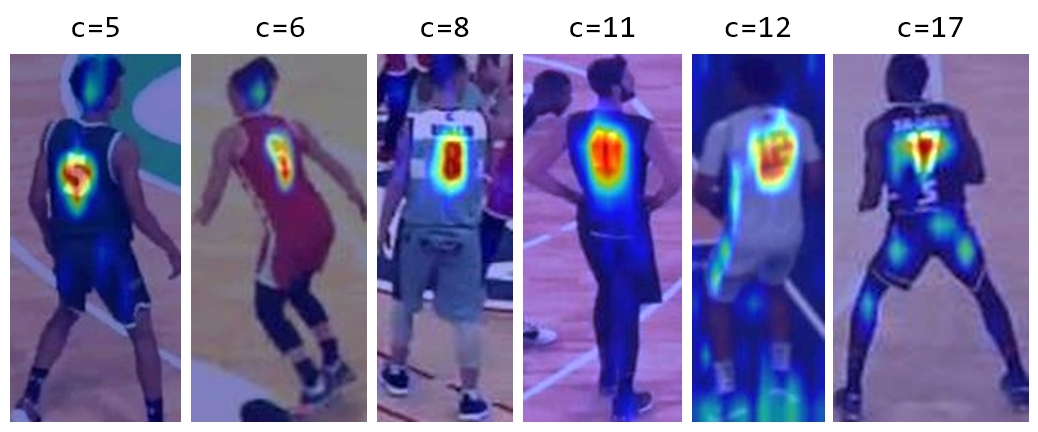}
	\caption{Jersey number localisation with Score-CAM.}
	\label{img:localisation_cam}
\end{figure}

With an appropriate text-prompt it is also possible to identify the image region where the jersey number is located. We use for the best results the text-prompt {\itshape "jersey number $\lbrace$c$\rbrace$, text number $\lbrace$c$\rbrace$"} with $c$ as the concrete number we are looking for. Figure~\ref{img:localisation_cam} shows the result of the Score-CAM algorithm visualising the image regions, that are most important, when calculating the similarity between the image and text-prompt with the concrete jersey number $c$. On most of the samples the image region with the jersey number is very well identified.

\subsubsection{Visualising Similarity with Score-Cam}

After fine-tuning of the model on the DeepSportRadar Player Re-identification dataset, zero-shot classification with text-prompts is no longer possible on CLIP models, because the text encoder does no longer match with the vision encoder. However, we can use the Score-CAM algorithm to visualise the most important image regions when calculating the cosine similarity between a query and a gallery image.

\begin{figure}[h]
\centering
	\includegraphics[width=0.45\textwidth]{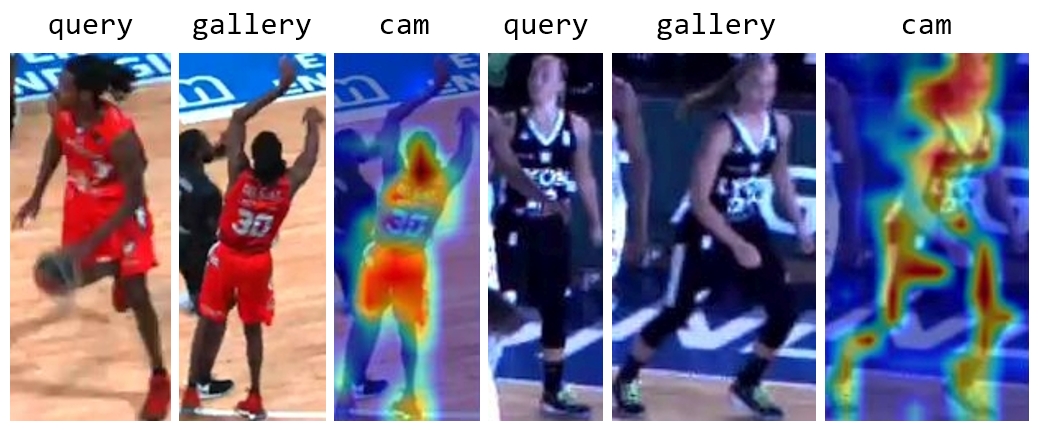}
	\caption{Similarity visualisation with Score-CAM after fine-tuning our model.}
	\label{img:similiarity_cam}
\end{figure}

On the left gallery image in Figure~\ref{img:similiarity_cam} the regions of the head, jersey and shoes are identified as most important. Here the jersey number is only visible on the gallery image. Therefore this region does not have a high similarity to the query image. 

In the right gallery image the pose of the player is different while large parts of the view remain unchanged compared to the query image. These results in a high similarity of the face and large parts of the body.

To summarise, contrastive learning leads to reasonable decisions about which image regions are relevant for a similarity matching task. The model is able to localise the image regions where the player is located very well and gets not distracted by the background or parts of other players on the image.

\section{Conclusion}

In this work, we investigate how to transfer the outstanding zero-shot performance of pre-trained CLIP models into the domain of player re-identification in sports. 
For our approach we reformulate the contrastive language-image pre-training from CLIP to a contrastive image-image training. On the MMSports 2022 Player Re-Identification challenge we achieve with a fine-tuned ViT-L/14 a mAP of 98.44\% and 97.56\% with a ViT-B/16. These scores are the results of single models without any ensemble. In our ablation study we apply the Score-CAM algorithm on our fine-tuned models. This visualisation leads to a better understanding which regions are most important when calculating the cosine similarity of a query and a gallery image. In addition we demonstrate that the CLIP pre-trained models have strong OCR capabilities to recognise and localise jersey numbers very well. Our work shows the superior performance of pre-trained CLIP models to combine multiple tasks such as jersey number detection, face recognition and colour matching to generate useful image representations for a re-identification task.

\section{Acknowledgement}
The authors gratefully acknowledge the computing time granted by the Institute for Distributed Intelligent Systems and provided on the GPU cluster Monacum One at the University of the Bundeswehr Munich.

{\small
\bibliographystyle{ieeetr}
\bibliography{egbib}
}
\end{document}